\documentclass{llncs}

\usepackage{latexsym}
\usepackage{amsfonts}
\usepackage[arrow,matrix,frame,curve,cmtip]{xy}\CompileMatrices

\makeatletter 
\g@addto@macro\@verbatim\scriptsize 
\makeatother

\title{Soft Constraint Logic Programming for Electric Vehicle Travel Optimization \thanks{
    Research partially supported by the EU FP7-ICT IP
    \textsc{ASCEns}.}  }

\author{
Giacoma Valentina Monreale\inst{1}, Ugo Montanari\inst{1} and Nicklas Hoch\inst{2}
\institute{Dipartimento di Informatica,
Universit\`a di Pisa, Italy
\\
vale@di.unipi.it, ugo@di.unipi.it
\and Volkswagen AG, Corporate Research Group
Wolfsburg, Germany \\ nicklas.hoch@volkswagen.de}
}
\titlerunning{SCLP for E-Vehicle Travel Optimization}
\authorrunning{Monreale and Montanari}

\begin{document}
\maketitle

\begin{abstract}
Soft Constraint Logic Programming is a natural and flexible declarative programming formalism,
which allows to model and solve real-life problems involving constraints of different types.

In this paper, after providing a slightly more general and elegant presentation of the framework,
we show how we can apply it 
to the e-mobility problem of coordinating electric vehicles in order to overcome both energetic and
temporal constraints and so to reduce their running cost.
In particular, we focus on the journey optimization sub-problem, 
considering sequences of trips from a user's appointment to another one.
Solutions provide the best alternatives in terms of time and energy consumption, including route 
sequences and possible charging events.

\end{abstract} 

\section{Introduction}

\emph{Classical constraint satisfaction problems} (CSPs) 
\cite{DBLP:books/daglib/0076790}
represent an expressive and natural formalism useful to specify different types of real-life problems.
A CSP can be described as a set of variables associated with a domain of values,
and a set of constraints. A constraint is a limitation of the possible combinations of the values of 
some variables. So, solving a CSP consists in finding an assignment of values
to all its variables guaranteeing that all constraints are satisfied.

Despite their applicability, the main limit suffered by CSPs is the ability of just stating if 
an assignment of certain values to the variables is allowed or not. This is indeed not enough
to model scenarios where the knowledge is not either entirely available or not crisp.
In these cases constraints are preferences and, when the problem is overconstrained, 
one would like to find a solution that is not so bad, i.e., the best solution
according to the levels of preferences.
For this reason, in \cite{DBLP:conf/ijcai/BistarelliMR95,DBLP:journals/jacm/BistarelliMR97}, 
the \emph{soft CSP} framework has been proposed.
It extends classical constraints by adding to the usual notion of CSP
the concept of a structure representing the levels of satisfiability or the costs of a constraint. 
Such a structure is represented by a semiring, that is, a set with two operations: 
one (usually denoted by $+$) is used to generate an
ordering over the levels, while the other one (denoted by $\times$) is used to define how two levels can be
combined and which level is the result of such a combination.

\emph{Constraint logic programming} (CLP) \cite{DBLP:conf/popl/JaffarL87} 
extends logic programming (LP) 
by embedding constraints in it: term equalities is replaced with constraints
and the basic
operation of LP languages, the unification, is replaced by constraint handling in a constraint system. 
It therefore inherits the
declarative approach of LP, according to which the programmer specifies what to compute
while disregarding how to compute it, by also offering efficient constraint-solving algorithms.

However, only classical constraints can be handled in the CLP framework.
So, in \cite{DBLP:journals/toplas/BistarelliR01}, it has been extended to also handle soft constraints. 
This has led
to a high-level and flexible declarative programming formalism, called \emph{Soft CLP} (SCLP),
allowing to easily model and solve real-life problems involving constraints of different types.
Roughly speaking, SCLP programs are logic programs where constraints are represented by predicates which are
defined by clauses whose body is a value of 
the semiring modelling the levels of satisfiability or the costs of the constraints.
The flexibility of the approach is due to the fact that the same framework can be used to handle different 
kinds of soft constraints by simply choosing different semirings.
It can indeed be used to handle fuzzy, probabilistic, prioritized and 
optimization problems, as well as classical constraints. 

In this paper, before presenting an application of the SCLP framework to the e-mobility, 
we provide 
a slightly more general and elegant presentation of the SCSP framework 
based on the notion of named semiring, as briefly presented in \cite{BuscemiM07}.
In particular, soft constraints are elements of the named semiring which, besides the additive and multiplicative 
operations useful to combine the constraints, is equipped by 
a permutation algebra and a hiding operator allowing an explicit handling of names.
The permutation algebra indeed allows to characterize the support
of each constraint, that is, the subset of variables on which the constraint really depends,
while the hiding operator allows us to remove
variables from the support of a constraint.
Then, the SCLP framework can be seen as the general case of the SCSP one.

We propose the SCLP framework 
as a high-level specification formalism useful to naturally model and also solve 
some e-mobility optimization
problems.
With e-mobility indeed 
new constraints must be considered: electric vehicles (EVs) have 
a limited range and they take long time to charge.
So, it is needed to guarantee that throughout the user's itinerary 
the EV never underruns a limit energy level and that the user always arrives in time to all
his appointments.

We consider the EV travel optimization problem described in \cite{VWPaper}.
In it the user has a set of appointments and he makes
a series of decisions about the sequences of trips from an appointment to
another one, for example which route to take, if and where to charge the EV and so on.
The aim is to find the optimal combination of travel choices which
minimizes the users cost criteria.
In \cite{VWPaper}, 
the authors propose a hierarchical presentation of the mobility
framework, which they exploit to decompose the optimization problem in 
sub-problems, and in particular, they study the journey level one. 
Here, beside it, we
consider the optimization problem of the lower level, i.e., the one of the trip level.
This last problem consists in finding the best trips
in terms of travel time and energy consumption, while
the former one consists in finding
the optimal
journey, that is, the optimal sequence of coupled trips, again in terms of the same criteria,
guaranteeing that the user reaches each appointment in time and that the state of
charge (SoC) of the EV never falls below a given threshold.

The trip level problem substantially coincides with the multicriteria version of the 
shortest path problem modelled in \cite{DBLP:journals/tocl/BistarelliMRS10} as an SCLP program. 
So, starting from a slightly different specification of this problem, we propose an SCLP program modelling
the journey problem.
In order to also actually execute both the SCLP programs, we propose
CIAO Prolog \cite{ciao-reference-manual-tr}, a system supporting constraint
logic programming. We therefore explicitly implement the soft framework, by
defining two predicates, the $plus$ and the $times$ ones,
which respectively model the additive and the multiplicative operations 
of the semiring.

This work arises from the research activity in the 
European project ASCENS (Autonomic Service Component ENSembles)
aiming at studying formal models, languages,
and programming tools, for the modelling and the development of 
autonomous, self-aware adaptive systems.
E-mobility is right one of the case studies of the project
and our work represents a first answer to the need of
a high-level, declarative, executable specification language and of a 
powerful and flexible programming environment where e-mobility problems
can be easily and naturally modelled and solved.
We indeed show as the SCLP framework is able to satisfy all these requirements 
and it can thus be used as a support for rapid prototyping and exploratory 
programming for this kind of problems.

The paper is organized as follows. Section \ref{sec:SCSP} introduces the SCSP framework 
as an instantiation of the
named semiring framework. Section \ref{sec:SCLP} briefly recalls the SCLP language and then Section \ref{sec:Emobility}
shows how the trip and journey optimization problems can be modelled and solved through (S)CLP programs.
Finally, Section \ref{sec:concl} concludes the paper by
illustrating some open venues for further works.

\section{Soft Constraints by means of Named Semirings} \label{sec:SCSP}

This section presents the soft CSP framework based on semiring 
\cite{DBLP:conf/ijcai/BistarelliMR95,DBLP:journals/jacm/BistarelliMR97}
as an instantiation of 
the more general framework based on named semiring \cite{BuscemiM07}.

The notion of named semiring is based on the ones of c-semiring (c stands for constraint) and permutation algebra.

\subsection{C-Semiring}
\begin{definition}[c-semiring]
A c-semiring is a tuple $\langle A, +, \times, 0,1 \rangle$ such that:
\begin{itemize}
\item $A$ is a set and $0, 1 \in A$;
\item $+: A \times A \to A$ is a commutative, associative and idempotent operation, such that $0$ is its unit element
and $1$ is its absorbing element;
\item $\times: A \times A \to A$ is a commutative and associative operation, such that it distributes over $+$, $1$
is its unit element, and $0$ is its absorbing element.
\end{itemize}
\end{definition}

Thanks to the idempotence of the
$+$ operator, the relation $\langle A, \leq \rangle$, defined as 
$a \leq b$ if $a + b = b$, is a partial order. Intuitively, $a \leq b$ means that 
$b$ is better than $a$ or that $a$ implies $b$. 

It is possible to prove that:
(i) the two operations $+$ and $\times$ are monotone on $\leq$; 
(ii) $0$ is its minimum and $1$ its maximum; (iii)$\langle A, \leq \rangle$ is a complete
lattice and $+$ is its least upper bound
\footnote{Actually, in order to prove this result, we must assume that the sum of an infinite number
of elements exists.}. 
Finally, if $\times$ is idempotent then: (iv)$+$ distribute over $\times$;
(v) $\langle A, \leq \rangle$ is a distributive lattice, and (vi) $\times$ is its greatest lower bound.

\subsection{Permutation Algebra}

Here we briefly recall the notion of \emph{permutation algebra}.
We refer the reader to \cite{DBLP:journals/lisp/GadducciMM06} for a detailed introduction.

In the following, we fix a chosen infinite, countable, totally ordered set $\mathcal N$ of names, which we denote 
by $x,y, z, \ldots$ .

\begin{definition}[Permutations]
A \emph{name substitution} is a
function $\sigma : \mathcal N \to \mathcal N$, while
a \emph{permutation} $\rho$ is a bijective name substitution.
The set of all such permutations on $\mathcal N$ is denoted by $\mathcal P(\mathcal N)$.
\end{definition}

\begin{definition}[Kernel]
Let 
$\rho \in \mathcal P(\mathcal N)$ be a permutation on $\mathcal N$.
The \emph{kernel} of $\rho$ is the set of the names that are
changed by the permutation, formally, $K(\rho) = \{ x \in \mathcal N | \rho(x) \not = x \}$. 

A permutation $\rho$ is \emph{finite} if its kernel is finite. 
\end{definition}

From now on we consider only finite permutations.

In the following, we introduce the notion of permutation algebra.
It consists of a pair composed of a carrier set and
a description of how the elements of the carrier set are transformed by permutations.

\begin{definition}[Permutation Algebras]
The \emph{permutation signature} $\Sigma_p$ on $\mathcal N$ is defined as
the set of unary operators $\{\widehat \rho | \rho \in \mathcal P (\mathcal N) \}$ plus
the two axioms $\widehat{id}(a) = a$ and $\widehat{\rho}_1(\widehat{\rho}_2(a)) = \widehat{\rho_1 \rho_2} (a)$. 

A \emph{permutation algebra} $\mathcal A = (A, \{\widehat \rho_A\})$ 
consists of a carrier set $A$ and the set of the interpreted operations $\widehat \rho_A$.

\end{definition}

\begin{definition}[Support]
Let $\mathcal A$ be a permutation algebra and $a$ an element of its carrier set.
The \emph{support} of $a$, $supp(a)$, is the smallest set of names such that, 
given a permutation $\rho$, if $\rho(x)=x$ for all $x \in supp(a)$, then $\widehat \rho_A(a)=a$.
\end{definition}

Intuitively, $supp(a)$ represents the free names of $a$: indeed the permutations which do
not modify them are not influent on $a$. 

\begin{definition}[Finitely supported algebra]
A permutation algebra $\mathcal A$ is
\emph{finitely-supported} if each element of its carrier has finite support. 
\end{definition}

\subsection{Named c-Semiring.}
A named semiring \cite{BuscemiM07} is a c-semiring
plus a finitely-supported permutation algebra $\mathcal A$ and a hiding operator $(\nu x.)$. 
The permutation algebra allows characterizing the finite set of free
names of each element $c$ of the named semiring (represented by the support
of $c$), 
while $(\nu x.)$ applied to $c$ makes 
the name $x$ local in $c$.

\begin{definition}[Fusion]
A (name) fusion is a total equivalence relation on $\mathcal N$ with only
finitely many non-singular equivalence classes. We denote by $x=y$ the fusion with a
unique non-singular equivalence class consisting of $x$ and $y$.
\end{definition}

\begin{definition}[Named c-semiring]
A named c-semiring $\mathcal C = \langle C,+,\times,\nu x.,\{\widehat \rho_C\},0,1 \rangle$ is a tuple where: 
\begin{itemize}
\item $x = y \in C$ for $x,y \in \mathcal N$; 
\item $\langle C,+,\times,0,1 \rangle$ is a c-semiring; 
\item $\langle C, \{\widehat \rho_C\} \rangle$ is a finite-support permutation algebra; 
\item $\nu x. : C \to C$, for each name $x$, is a unary operation; 
\item for all $c,d \in C$ and for all $\rho$ the following axioms hold:
\end{itemize}
\begin{tabular}{p{1.5cm}p{12cm}}
(FUSE) & $x=y \times c \,$ iff $\, x=y\times [y/x]c$\\
(HIDE) & $\nu x. 1 = 1$ \hspace{3mm} $\nu x.\nu y.c = \nu y.\nu x.c$ \hspace{3mm} 
$\nu x.(c \times d) = c\times \nu x.d$ if $x \not \in supp(c)$\\
& $\nu x. (c+d) = c+ \nu x.d$ if $x \not \in supp(c)$ \hspace{3mm} $\nu x. c = \nu y. [y/x]c$ if $y \not \in supp(c)$\\
(PERM) & $\widehat \rho_C 0 = 0$ \hspace{3mm} $\widehat \rho_C 1 = 1$ \hspace{3mm} 
$\widehat \rho_C (c\times d) = \widehat \rho_C c \times \widehat \rho_C d$ \hspace{3mm} 
$\widehat \rho_C (c+d) = \widehat \rho_C c + \widehat \rho_C d$\\
&$\widehat \rho_C(\nu x.c) = \nu x. (\widehat \rho_C c)$ if $x \not \in K(\rho)$
\end{tabular}
\end{definition}

In the (FUSE) axiom, $[y/x]c$ denotes $c$ where $y$ is replaced by $x$.

\subsection{The Named SCSP Framework}
As briefly shown in \cite{BuscemiM07}, named c-semirings can be suitably instantiated to model 
SCSPs. 

\begin{definition} [Constraints]
Let $S = \langle A, +, \times, 0,1 \rangle$ be a c-semiring, 
$\mathcal N$ a set of totally ordered names, and $D$ a finite domain of interpretation for $\mathcal N$.
A \emph{soft constraint} is a function $(\mathcal N \to D) \to A$, which associates a value of $A$ to each
assignment $\eta: \mathcal N \to D$ of the names.

We define $\mathcal C$ as the set of all soft constraints over $\mathcal N$, $D$ and $A$.
\end{definition}

\begin{definition}
Let $S = \langle A, +, \times, 0,1 \rangle$ be a c-semiring, 
$\mathcal N$ a set of totally ordered names, and $D$ a finite domain of interpretation for $\mathcal N$.
Moreover, let $\mathcal C$ be the set of all soft constraints over $\mathcal N$, $D$ and $A$.
We define the $\mathcal C_{SCSP}$ as the named c-semiring 
$\langle \mathcal C,+',\otimes,\nu x., \{\widehat \rho_{\mathcal C}\},0',1' \rangle$,
where fusions $x=y$ are defined as $(x=y)\eta = 1$ if $\eta(x) = \eta(y)$ and $(x=y)\eta = 0$ otherwise;
$(c_1 +' c_2) \eta = c_1 \eta + c_2 \eta$; $(c_1 \otimes c_2) \eta = c_1 \eta \times c_2 \eta$;
$(\nu x.c)\eta= \sum_{d \in D}(c\eta [d/x])$; $(\widehat \rho_{\mathcal C} c)\eta = c \eta'$ with $\eta'(x)= \eta(\rho(x))$;
$0' \eta = 0$ and $1' \eta = 1'$ for all $\eta$.
\end{definition}

The assignment $\eta [d/x]$ is defined as usual: $\eta [d/x](y) = d$ if $x = y$ and 
$\eta [d/x](y)= \eta(y)$ otherwise.

Note that, by definition, each constraint $c \in \mathcal C$ involves all the
names in $\mathcal N$, but it really depends on the assignment of the names in $supp(c)$,
and intuitively, restricting means eliminating a name from the support of the constraint,
by choosing the best value for it.

\begin{definition}[Soft CSP]
A soft constraint satisfaction problem (SCSP) is a pair $\langle C, Y \rangle$, where
$C \subseteq \mathcal C$ is a set of constraints and $Y \subseteq \mathcal N$ is a set of names.
\end{definition}

Intuitively, $Y$ represents the set of interface names of the constraint set $C$.

\begin{definition}[Solution] 
Let $P = \langle C, Y \rangle$ be an SCSP. The solution of $P$ is the
constraint $Sol(P) = (\nu x_1) \ldots (\nu x_n) (\bigotimes C)$, 
with $\{ x_1, \ldots ,x_n\} = \bigcup_{c_i \in C} supp(c_i) \setminus Y$.
\end{definition}

\begin{definition}[Best level] 
Let $P = \langle C, Y \rangle$ be an SCSP. Then,
the best level of consistency of $P$ is defined as the constraint 
$blevel(P) = (\nu x_1) \ldots (\nu x_n) (\bigotimes C)$, where $\{x_1,\ldots ,x_n\} = \bigcup_{c_i \in C} supp(c_i)$.
\end{definition}

Varying the semiring $S$, on which 
the named semiring $\mathcal C_{SCSP}$ is based,
several kinds of problems can be represented:
we consider the semiring $S_{CSP} = \langle \{true, false\}, \lor, \land, false, true \rangle$ for classical CSPs;
$S_{FCSP} = \langle \{x | x \in [0,1]\}, max, min, 0, 1 \rangle$ for fuzzy CSPs;
and $S_{WCSP} = \langle \mathbb N \cup \{+\infty\}, min, +, +\infty, 0 \rangle$ for optimization CSPs.

\begin{example} \label{ex:SCSP}
Let $P$ be the SCSP with three names $\mathcal N = \{x, y, z\}$, which can take values in
$D = \{red, blue, green\}$.  
We assume that only $x$ and $y$ are of interest, while
the 
constraints model the fact that all 
names must take different values.

We consider the semiring $S_{WCSP}$, introduced above, and we define 
three constraints, one for each pair of names, $c_{xy}$, $c_{yz}$ and $c_{zx}$,
associating the worst semiring value to the assignments
which give the same color to both the names of interest for the constraint.
In particular, we associate $+\infty$ to all the assignments giving the same color to both names,
if at most one name takes the color red then we associate $1$, otherwise we associate $2$.

In Horn logic this problem can be expressed as the predicate $P(x,y)$ below: \\
\indent
$P(x,y):- Q(x,y),Q(y,z),Q(z,x)$ \\ \indent
$Q(v,w):-$ if $v = w$ then $+\infty$ else if either $v$ or $w$ are red then $1$ else $2$.
\noindent

The predicate $Q(v,w)$ represents the constraint shown on the left of Fig. \ref{fig:SCSP},
where the names $v$ and $w$ represents the only names of the support.
Note that in this representation we only show the assignment for the names of the support of the constraint.
Therefore, each table entry actually represents different entries, one for each possible color of the 
names which are not in the support.
The constraint representing the problem is then represented by the combination of three constraints.
Indeed, $Q(x,y)$, $Q(y,z)$ and $Q(z,x)$ respectively represent the three constraints
$c_{xy}$, $c_{yz}$ and $c_{zx}$,
obtained by applying to the constraint represented by $Q(v,w)$ the permutations
mapping the names of the support $v$ and $w$ to the names of interest of the three constraints.
As said above, the relevant names of the problem are just $x$ and $y$, thus
the solution of the problem is obtained by combining the three constraints and restricting the scope
of the name $z$. It can be expressed as
$(\nu z) (c_{xy} \otimes c_{yz} \otimes c_{zx})$.
The resulting constraint, represented by the predicate $P(x,y)$, is shown on the right of Fig. \ref{fig:SCSP}.
Also in this case, we show the assignment of just the two names of the support.
Each table entry is the minimum value among the ones of the solutions providing a different color for $z$.  
In this case, the best level of solution is $4$, the minimum over all the entries.

\begin{figure}[t]
\center
\begin{tabular}{cc}
\begin{tabular}{l|l|c}
x & y &  \\
\hline 
red & red & $+\infty$ \\
red & blue & $1$ \\
red & green & $1$ \\
blue & blue & $+\infty$ \\
blue & red & $1$ \\
blue & green & $2$ \\
green & green & $+\infty$ \\
green & red & $1$ \\
green & blue & $2$ \\
\hline 
\end{tabular}
& \hspace{2cm}
\begin{tabular}{l|l|c}
v & w &  \\
\hline 
red & red & $+\infty$ \\
red & blue & $4$ \\
red & green & $4$ \\
blue & blue & $+\infty$ \\
blue & red & $4$ \\
blue & green & $4$ \\
green & green & $+\infty$ \\
green & red & $4$ \\
green & blue & $4$ \\
\hline 
\end{tabular}
\\ \\
\hspace{5mm} $Q(v, w)$ & \hspace{2cm} $Sol(P)$ 
\end{tabular}
\caption{The constraint represented by $Q(v, w)$ and the solution of the SCSP $P$.} \label{fig:SCSP}
\end{figure}

\end{example}

\section{Soft Constraint Logic Programming} \label{sec:SCLP}
This section briefly introduces the \emph{soft constraint logic programming} (SCLP).
For a more detailed and complete introduction, we refer the reader to \cite{DBLP:journals/toplas/BistarelliR01}.

The SCLP framework extends the classical constraint logic programming to also handle 
SCSPs. We can say that an SCLP program over a certain c-semiring $S$ is just a CLP program where constraints
are defined over $S$.
In the following, we fix a semiring $S = \langle A, +, \times, 0,1 \rangle$.

An SCLP program is hence a set of clauses composed of a head and a body, plus a goal. 
The head of a clause is simply an atom, while the body can be either a collection of atoms, or
a c-semiring value, or a special symbol 
$\Box$, denoting that it is empty.
In this two last cases clauses are called facts and define predicates representing constraints.
When the body is empty, we interpret it as $1$, the best element of the semiring. 
Atoms are $n$-ary predicate symbols followed by a tuple of $n$ terms, which can be
either a constant or a variable or an n-ary function symbol followed by $n$ terms.
Ground terms are terms without variables, and finally, a goal is a collection of atoms. 

\begin{example} \label{ex:SCLP}
As an example, consider the simple SCLP program on the left of Fig. \ref{fig:SCLP}, previously proposed in 
\cite{DBLP:journals/toplas/BistarelliR01}.
We consider the semiring $S_{KCSP}$ and the domain $D=\{a,b,c\}$.
The program is composed of six clauses. The last two are facts and
the semiring values $2$ and $3$, associated respectively with
the atoms $t(a)$ and $r(a)$, mean that they respectively cost $2$ and $3$ units.
The set $\mathcal N$ of the semiring contains all possible
costs, and the operations $min$ and $+$ allows us to minimize
the sum of the costs.
We consider as goal the atom \verb|:-s(a)|: later on we will show its semantics.
\end{example}

\begin{figure}[t]
\center
\begin{tabular}{cc}
\begin{tabular}{ll}
\verb|s(X)| & \verb|:- p(X,Y).| \\
\verb|p(a,b)| & \verb|:- q(a).| \\
\verb|p(a,c)| & \verb|:- r(a).| \\
\verb|q(a)| & \verb|:- t(a).| \\
\verb|t(a)| & \verb|:- 2.| \\
\verb|r(a)| & \verb|:- 3.| \\
\end{tabular}
&
\hspace{1.5 cm}
\begin{tabular}{l|ccccc}
& $I_1$ & $I_2$ & $I_3$ & $I_4$ \\
\hline
t(a) & 2 & 2 & 2 & 2  \\
r(a) & 3 & 3 & 3 & 3  \\
q(a) & $+\infty$ & 2 & 2 & 2 \\
p(a,c) & $+\infty$ & 3 & 3 & 3 \\
p(a,b) & $+\infty$ & $+\infty$ & 2 & 2 \\
s(a) &  $+\infty$ & $+\infty$ & 3 & 2 \\
s(b) &  $+\infty$ & $+\infty$ & $+\infty$ & $+\infty$ \\
s(c) &  $+\infty$ & $+\infty$ & $+\infty$ & $+\infty$ 
\end{tabular}
\end{tabular}
\caption{An example of SCLP program and its fix point semantics.} \label{fig:SCLP}
\end{figure}

Three equivalent semantics for the SCLP languages have been defined in \cite{DBLP:journals/toplas/BistarelliR01}: 
the model-theoretic, the fix-point, and the operational one. These
semantics are conservative extensions of the corresponding ones for logic programming:
this means that by choosing the c-semiring $S_{CSP}$ we get exactly the LP semantics.

Actually, we can see the SCLP framework as the general case of the SCSP one.
As shown in Example \ref{ex:SCSP}, we can indeed express an SCSP program as a set of
predicates representing the constraints
together with a unique clause which represents the problem and combines all constraints.
In the case of the SCLP, the main difference is that the depth of clause nesting is unlimited and 
the possible values $D$ of variables are 
elements of the Herbrand universe.
So, the meaning of a predicate $P(x,y)$ assigns a semiring value to all evaluations of variables $x$ and $y$ 
to the Herbrand domain.
This is exactly what for example the fix-point semantics of the SCLP language does.
%
%

In order to present the fix-point semantics, we need to introduce the
notion of interpretation and the $T_P$ operator, mapping interpretations into interpretations.

\begin{definition}[Interpretation]
%
An \emph{interpretation} $I$ consists of a domain $D$, representing the Herbrand universe, 
together with
a function which takes a predicate and an instantiation of its arguments (that is, a
ground atom), and returns an element of the semiring: 
$I: \bigcup _n (P_n \to (D^n \to A))$, where $P_n$ represents the set of n-ary predicates and 
$A$ is the set of the values of the semiring.
\end{definition}
%
%
%

Since interpretations are functions from ground
atoms to semiring values, we consider programs
composed of clauses where the head and the body contain only ground atoms.
So, for example, in the SCLP program of Example \ref{ex:SCLP}, 
the clause \verb|s(X) :- p(X,Y)| is replaced with all its instantiations.
In particular, for each $d \in D$, we have three clauses \verb|s(d) :- p(d,a)|,
\verb|s(d) :- p(d,b)|, and \verb|s(d) :- p(d,c)|.

\begin{definition}[$T_P$ Operator] 
Let $P$ be an SCLP program and $IS_P$ the set of all its interpretations.
Moreover let $I$ be an interpretation and $GA$ a ground atom,
such that $P$ contains $k$ clauses defining the predicate in $GA$
and the clause $i$ is of the shape $GA :- B^i_1, \ldots , B^i_{n_i}$. 
Then, we define the operator 
$T_P : IS_P \to IS_P$ as $T_P(I)(GA) = \sum^k_{i=1}(\prod_{j=1}^{n_i} I(B_j^i))$.
Whenever $B^i_j$ is a semiring value, its meaning $I(B^i_j)$ is fixed in any interpretation 
$I$ and it is the semiring value itself.
\end{definition}

\begin{definition}[Partial Order of Interpretations] 
Let $P$ be a program and $IS_P$ the set of all its interpretations.
We define the structure $\langle IS_P , \preceq \rangle$, where
$\forall I_1, I_2 \in IS_P, I_1 \preceq I_2$ if $I_1(GA) \leq I_2(GA)$ for any ground atom $GA$, 
where $\leq$ is the order induced by the semiring.
\end{definition}

Note that $\langle IS_P, \preceq \rangle$ is a complete partial order and its glb 
coincides with the glb operation in the lattice $A$ (extended to interpretations).
Since the function $T_P$ is monotone and continuous over this complete partial order,
then $T_P$ has a least fix-point $lfp(T_P) = glb(\{I| T_P(I) \preceq I\})$
and it can be obtained 
by computing $T_P \uparrow \omega$, i.e., by applying
$T_P$ to the bottom of the partial order of interpretations, and then 
repeatedly applying it until a fix-point.

\begin{example}
Consider again the SCLP program in Fig. \ref{fig:SCLP}.
In the definition of the $T_P$ operator we have to consider the additive and multiplicative operations
of the semiring $S_{KCSP}$, that is, the $min$ and $+$ operations.
As in \cite{DBLP:journals/toplas/BistarelliR01}, 
we start the computing of the semantics from the bottom of the partial order of interpretations, $I_0$,
which maps each semiring element into itself and each ground atom into $+\infty$.
%
The table on the right of Fig. \ref{fig:SCLP} shows the value associated by the interpretations 
to the most interesting ground atoms.
The interpretation $I_4$ represents the fix-point of $T_P$.
As an example, we show the computation for one of the most interesting case, the ground atom \verb|s(a)|,
which also corresponds to our goal.
We said that the clause \verb|s(X) :- p(X,Y)| is considered equivalent to all its instantiations.
Therefore, $I_4(s(a))= min\{I_3(p(a,a)),I_3(p(a,b)),I_3(p(a,c))\} = min \{+\infty, 2, 3\} = 2$. 
\end{example}
\section{The Electric Vehicle Travel Optimization Problem}
\label{sec:Emobility}

This section presents the EV travel optimization problem, introduced in \cite{VWPaper},
and shows how it can be naturally modelled and solved in the SCLP framework.

\paragraph{\bf{General description of the problem.}}
A user has a set of appointments, each of them is in a location
and has a starting time and a duration. The user makes a series of decisions
regarding the sequences of trips from an appointment to another one.
For example, he decides which route he wants to follow, where to park and if and how
to charge the EV at the appointment location.

All possible combinations of travel choices form the choice set.
A travel choice is optimal if 
it minimizes the user's cost criteria.

In particular, finding a single optimal trip consists in finding the best trips in terms of
travel time and energy consumption. 
Finding instead the optimal journey, that is, the optimal sequence of coupled trips, 
consists in finding the best journeys not only in terms of travel time and
energy consumption, but also in terms of other important criteria for the user, such as the
charging cost, the number of charging events, etc.
However, in finding the solution it needs to guarantee that the user reaches 
each appointment in time and that the state of charge (SoC) of the
vehicle never falls below a predefined threshold. 

In \cite{VWPaper}, the authors propose a hierarchical presentation of the e-mobility framework, 
which they exploit to decompose the optimization problem in sub-optimization problems.
In particular, they identify four levels of mobility: the \emph{component level}, whose main tasks are
the inter- and intra-component coordination; the \emph{trip level},
whose main task is the time and energy optimal routing;
the \emph{journey level}, which handles sequences of trips together with charging and parking strategies; 
and the \emph{mobility level}, which handles mobility services, such as, car and ride service.

Each level represents a different optimization problem and the results of the lower level
will be inputs of the higher level. However, since, in general, the best solution of the lower level
could be not optimal for the higher one, the results of the lower level could contains
several solutions and not only the best one. 

In the following, we consider only the trip level and the journey level optimization problems.
In particular, we first present their formalization and then we show how we can model them in the SCLP framework.

\paragraph{\bf{Formalizations of the trip and journey level optimization problems.}} 

The trip level optimization problem substantially coincides with the 
multi-criteria shortest path problem.
The road network is indeed represented by a directed graph $G :=(N,E)$,
where each arc $e \in E$ from a node $p$ to a node $q$ has associated a label $\langle c_T, c_E \rangle$, that is, a pair
whose elements represent the costs, respectively in terms of time and energy consumption,
of the arc from $p$ to $q$.

\begin{figure}[t]
\center
\begin{tabular}{ccc}
\begin{tabular}{c}
\xymatrix@C=40pt@R=40pt{
& *++[o][F-]{r} \ar[r]^{\langle 3,3 \rangle} \ar@/^/[d]^{\langle 1,1 \rangle} & *++[o][F-]{s} \ar[d]^{\langle 1,1 \rangle} \\
*++[o][F-]{p} \ar@/_1pc/[rr]_{\langle 3,9 \rangle} \ar[ur]^{\langle 2,7 \rangle} \ar[r]^{\langle 2,4 \rangle} & 
*++[o][F-]{q} \ar@/^/[u]^{\langle 1,1 \rangle} \ar[ru]_{\langle 4,8 \rangle} \ar[r]^{\langle 2,4 \rangle} &
 *++[o][F-]{t} 
}
\end{tabular}
&
\begin{tabular}{c}
\begin{tabular}{c|c|c}
Loc. & Start. time & Dur. \\
\hline
p & 7 & 1 \\
r & 11 & 2 \\
t & 18 & 3
\end{tabular}
\end{tabular}
&
\begin{tabular}{c}
\begin{tabular}{c|c|c}
Name & Spots & Loc. \\
\hline
csp1 & 7 & p \\
csr1 & 4 & r \\
csr2 & 0 & r
\end{tabular}
\end{tabular} \\
Road Network & Appointments & Charging Stations \\
\end{tabular}
\caption{The road network, the user's appointments and the charging stations.} \label{fig:graph}
\end{figure}
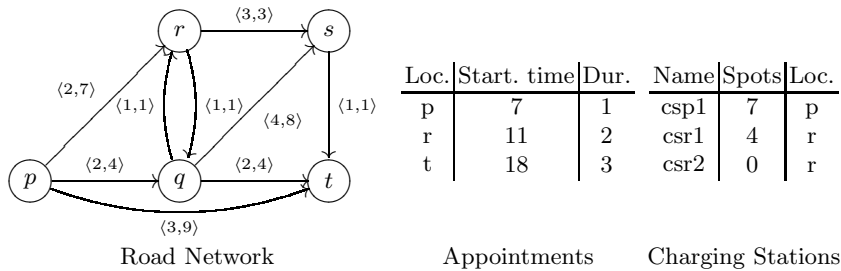

So, given the road network $G$, such as the one on the left of Fig. \ref{fig:graph}, 
a source node $n_s$ and a destination node
$n_d$, the problem consists in finding all the best paths between $n_s$ and $n_d$ in terms of time and energy consumption.
Note that, since the costs of the arcs are elements of a partially ordered set, the solution 
can contain several paths, 
that is, all paths which are not dominated by others, but which have
different incomparable costs. For example, if we 
want to know the best paths from $p$ to $t$ in the graph of Fig. \ref{fig:graph}, 
the solution will contain both the paths $\{p,t\}$
with cost $\langle 3,9\rangle$ and $\{p,q,t\}$ with cost $\langle 4,8\rangle$. The former is indeed better in terms of time
while the latter in terms of energy consumption.

As far as the journey level optimization problem is concerned, we use the formalization 
presented in \cite{VWPaper}. Actually, we consider a simpler version of it, which avoids to consider
car parks and the time that the users would take to go from either the car park or the charging station
to the location of the appointment.
Moreover, we consider only the time and energy consumption as cost criteria to be minimized.
All these simplifications allows a slender and more readable presentation of the SCLP program modelling the problem.

Let $A = \{A_1, \ldots, A_n\}$ be the set of the user's appointments.
In order to describe the problem, we use different time variables.
All of them have the shape $_it_Z^Y$, where 
$i$ denotes the appointment,
$Y \in \{D,A\}$ ($D$ stands for drive and $A$ for appointment), and
$Z \in \{S,E\}$ ($S$ stands for start and $E$ for end).

Each appointments is defined by a location $L_i$, a starting time $_it^A_S$, an end time $_it^A_E$ and therefore a 
duration $_id^A$.
In order to go from an appointment $A_i$ to the next one $A_{i+1}$, the user 
leaves with an
EV from the location $L_i$ at time $_it_S^D$ and
drive to location $L_{i+1}$. The user travels along the route alternative $iR^D$ (computed by the trip level problem),
which consumes energy and hence reduces the SoC. 
Obviously, the chosen route must allow the user to arrive to destination with the SoC of his EV.
We assume that the SoC always decreases
during driving and increases during charging events\footnote{This 
is the typical behaviour of EVs, however, as explained in \cite{VWPaper}, 
in particular cases it might also increase during driving
and decreases during charging.}.
The user
arrives at $_it_E^D$ and the appointment starts at $_it^A_S$.
The user must arrive in time to the appointment, so it is required that $_it_E^D \leq \, _it^A_S$.
During the appointment, it is also possible to schedule a charging event if the SoC of the
EV is not enough to continue the journey. We assume to have
a set of charging stations. Each of them is simply defined 
by its name $CSname$, the number of available charging spots $SpotsNum$, and the location $L$ where it is.

Therefore, given the road network $G$,
a set of appointments, as the ones described in the table in the middle of Fig. \ref{fig:graph}, 
and a set of charging stations, as the ones in the rightmost table of Fig. \ref{fig:graph},
the problem consists in finding all the best journeys through all the appointment locations 
in terms of time and energy consumption.
As for the travel optimization problem, also here the solution 
can contain several journeys,
that is, all the non-dominated ones.

\paragraph{\bf{SCLP programs for the optimization problems.}}

In the following, we show how the
SCLP framework can be used as a linguistic support and a high-level and flexible
programming environment where naturally modeling and solving the two optimization problems
presented above.

As far as the trip level optimization problem, we propose a slightly different version of the
model proposed in \cite{DBLP:journals/tocl/BistarelliMRS10} for the multi-criteria shortest path problem.
So, as there, we consider an SCLP program over the
c-semiring denoted $P^H(S)$ which, given a source node $n_s$ and a target node $n_d$,
allows us to obtain the set of the costs of
all non-dominated paths from $n_s$ to $n_d$. 

The semiring $P^H(S)$ is obtained starting from a semiring $S = \langle A, +, \times, 0,1 \rangle$,
which in our case is the one modelling the costs associated to each edge, i.e, 
$S = \langle N^2, min', +', \langle \infty, \infty \rangle , \langle 0,0 \rangle \rangle$,
where $min'$ and $+'$ are the $min$ and $+$ operations extended to pairs.
Indeed, in general we want to minimize the sum of each cost, but, since we want to obtain all the non dominated paths, 
we 
consider 
$P^H(S)$.

Given a semiring $S$, we define $P^H(S) = \langle P^H(A), \uplus,\times^*, \emptyset, A \rangle$, where
$P^H(A)$ is the Hoare Power Domain of $A$, that is, $P^H(A)=\{S \subseteq A | x \in S, y \leq_S x$ implies $y \in S\}$.
These sets are isomorphic to those
containing just the non-dominated values, thus, in the following, we will use this more
compact and efficient representation, where each element
of $P^H(A)$ will represent the costs of all non-dominated paths from a node to another one.
The top element of the semiring is the set $A$ (its compact form is \{1\}, which in our example
is $\{\langle 0, 0	\rangle\})$; the bottom element is the empty set;
$\uplus$ is the formal union that takes two sets and gives
their union; $\times^*$ takes two sets and produces another one obtained by multiplying
(using the multiplicative operation of the original semiring, in our case $+'$)
each element of the first set with each element of the second one.

Note that, in the partial order induced by the additive operation of this semiring, 
$a \leq_{P^H(S)} b$ intuitively means that for each
element of $a$, there exists an element of $b$ which dominates it (in the partial order
of the original semiring).

Following \cite{DBLP:journals/tocl/BistarelliMRS10}, in order to also really execute the SCLP program,
we model the problem with a program in CIAO Prolog 
\cite{ciao-reference-manual-tr}, a system supporting CLP,
by explicitly implementing the soft framework. 
The program is shown in Fig. \ref{fig:CiaoPaths}.

\begin{figure}[t]
\center
\begin{tabular}{p{5.5cm} p{6cm}}
\begin{verbatim}
:-module(paths,_,_).
:-use_module(library(lists)).
:-use_module(library(aggregates)).

minPair([T,E],[T1,E1]):-
  T < T1,
  E < E1.

times([T1,E1],[T2,E2],[T3,E3]):-
  T3 = T1 + T2, 
  E3 = E1 + E2.

plus([],L,[]).
plus([[P,T,E]|RestL],L, 
  [[P,T,E]|BestPaths]):-
  nondominated([T,E],L),
  plus(RestL,L,BestPaths).
plus([[P,T,E]|RestL],L,BestPaths):-
  \+nondominated([T,E],L),
  plus(RestL,L,BestPaths).

nondominated([T,E],[]).
nondominated([T,E],[[P,T1,E1]|L]):-
  \+minPair([T1,E1],[T,E]),
  nondominated([T,E],L).
\end{verbatim}
&
\begin{verbatim}




edge(p,q,[2,4]).  edge(q,t,[2,4]).
edge(p,r,[2,7]).  edge(r,s,[3,3]).
edge(p,t,[3,9]).  edge(r,q,[1,1]).
edge(q,r,[1,1]).  edge(s,t,[1,1]).
edge(q,s,[4,8]).
			
path(X,Y,[X,Y],_,[T,E],Lim):-
  edge(X,Y,[T,E]),
  E =< Lim.

path(X,Y,[X|L],V,[T,E],Lim):-
  edge(X,Z,[T1,E1]),
  nocontainsx(V,Z),
  path(Z,Y,L,[Z|V],[T2,E2],Lim),
  times([T1,E1],[T2,E2],[T,E]),
  E =< Lim.

paths(X,Y,Lim,BestPaths):-
  findall([P,T,E],path(X,Y,P,[X],[T,E],Lim),ResL),
  plus(ResL,ResL,BestPaths).
\end{verbatim}
\end{tabular}
\vspace*{-5mm}
\caption{The CIAO program modelling the trip level optimization problem.} \label{fig:CiaoPaths}
\end{figure}

Here we consider the road network presented in Fig. \ref{fig:graph}, so we have a set of clauses
modelling it. In particular, we have a set of facts modelling all the edges of the graph. Each fact 
has the shape $edge(n_s,n_d, [c_T,c_E])$,
where $n_s$ represents the source node, $n_d$ represents the destination node and the pair
$[c_T,c_E]$ represents the costs of the edge in term of time and energy.
Note that, differently from what would happen in the pure SCLP framework, 
these facts (representing constraints) have the cost in the head of the clauses
and not in the body. This is needed for implementing the soft framework, and in particular
the two operations of the semiring.

Moreover, there are two clauses $path$ describing the structure of paths:
the upper one models the base case, where
a path is simply an edge, while the lower one represents the recursive case, 
where a path is an edge plus another path.
The head of the path clauses has the following shape
$path(n_s,n_d,L_N,L_V,[c_T,c_E],Lim)$, where
$n_s$ and $n_d$ are respectively the source and destination nodes,
$L_N$ is the list needed to remember, at the end, all the visited
nodes of the path in the ordering of the visit,
$L_V$ is the list of the already visited nodes needed to avoid infinite recursion
where there are graph loops,
$[c_T,c_E]$ is used to remember the cost of the path in terms of time and energy, and 
finally, $Lim$ represents the maximum amount of energy that the EV can consume. It is used to retrieve only
the paths with a total cost in terms of energy equal to or less than the passed value. 

The $times$ and $plus$ clauses are useful to model the soft framework. In particular,
the first clause is useful to model the multiplicative operation of the semiring allowing us
to compose the global costs of the edges
together, time with time and energy with energy.
The $plus$ predicate instead mimics the additive operation and it is useful to find the 
best, i.e. non-dominated, paths among all the possible solutions.
The $plus$ predicate is indeed used in the body of the $paths$ clause, which
collects all
the paths from a given source node to a given destination node and 
returns the best solutions chosen with the help
of the $plus$ predicate.
So, if we want to know the best paths, in the graph of Fig. \ref{fig:graph}, 
from $p$ to $t$ 
with a total cost in terms of energy consumption less than or equal to $10$,
we have to perform the CIAO query $paths(p, t, 10, BestPaths)$,
where the $BestPaths$ variable will be instantiated with the 
list containing all the non-dominated paths. In particular, for each of them,
the list will contain the sequence of the
nodes in the path and the total cost of the
path in terms of time and energy.
The output of the CIAO program for this query is shown in Fig. \ref{fig:pathsRes}.

\begin{figure}[t]
\begin{center}
\begin{verbatim}
Ciao 1.14.2-13646: Mon Aug 15 10:49:59 2011

?-   paths(p,t,10,BestPaths).

BestPaths = [[[p,t],3,9],[[p,q,t],2+2,4+4]] ?.
no
?-
\end{verbatim}
\end{center}
\vspace*{-5mm}
\caption{The output for the query $paths(p,t,10,BestPaths)$.} \label{fig:pathsRes}
\end{figure}

Now, by using the SCLP program modelling the travel optimization problem, we can also show 
the one modelling the journey level problem. Also in this case, as before, we 
consider the $P^H(S)$ semiring and we propose a CIAO program,
where we also model the soft framework.
%
%
The CIAO program modelling the journey optimization problem is presented in Fig. \ref{fig:CiaoJourneys}.

\begin{figure}[t]
\center
\begin{tabular}{p{4.7cm} p{6cm}}
\begin{verbatim}
:-module(journey,_,_).
:-use_module(paths).

plus([],L,[]).
plus([[P,T,E,ChEv]|RestL],L,
      [[P,T,E,ChEv]|BestPaths]):-
  nondominated([P,T,E],L),
  plus(RestL,L,BestPaths).
plus([[P,T,E,ChEv]|RestL],L,
                     BestPaths):-
  \+nondominated([P,T,E],L),
  plus(RestL,L,BestPaths).


nondominated([P,T,E],[]).
nondominated([P,T,E],
          [[P1,T1,E1,ChEv1]|L]):-
  \+minPair([T1,E1],[T,E]),
  nondominated([P,T,E], L).


appointment(p,7,1).
appointment(r,11,2).
appointment(t,18,3).

chargingStation(csp1,7,p).
chargingStation(csr1,4,r).
chargingStation(csr2,0,r).


journeys(Places,EV,BestJourneies):-
findall([P,T,E,ChEv],journey(Places,P,ChEv,[T,E],SoC),ResL),
plus(ResL,ResL,BestJourneies).

\end{verbatim}
&
\begin{verbatim}
journey([X,Y],[P],[],[T,E],SoC):-
  appointment(X,Tx,Dx), appointment(Y,Ty,Dy),
  path(X,Y,P,[X],[T,E],SoC),
  timeSum(Tx,Dx,T,ArrT), ArrT=<Ty.

journey([X,Y],[P],[[X,ID]],[T,E],SoC):-
  appointment(X,Tx,Dx), appointment(Y,Ty,Dy),
  \+path(X,Y,P,[X],[T,E],SoC),
  chargingStation(ID,Spots,X),Spots>0,
  newSoC(SoC,Dx,NewSoC),
  path(X,Y,P,[X],[T,E],NewSoC),
  timeSum(Tx,Dx,T,ArrT), ArrT=<Ty.
  
journey([X|[Y|Z]],[P|LP],ChEv,[T,E],SoC):-
  appointment(X,Tx,Dx), appointment(Y,Ty,Dy),
  path(X,Y,P,[X],[T1,E1],SoC),
  timeSum(Tx,Dx,T1,ArrT), ArrT=<Ty,
  journey([Y|Z],LP,ChEv,[T2,E2],(SoC-E1)),
  times([T1,E1],[T2,E2],[T,E]).

journey([X|[Y|Z]],[P|LP],[[X,ID]|ChEv],[T,E],SoC):-
  appointment(X,Tx,Dx), appointment(Y,Ty,Dy),
  \+path(X,Y,P,[X],[T1,E1],SoC),
  chargingStation(ID,Spots,X), Spots>0,
  newSoC(SoC,Dx,NewSoC),
  path(X,Y,P,[X],[T1,E1],NewSoC),
  timeSum(Tx,Dx,T1,ArrT), ArrT=<Ty,
  journey([Y|Z],LP,ChEv,[T2,E2],(NewSoC-E1)),
  times([T1,E1],[T2,E2],[T,E]).
\end{verbatim}
\end{tabular}
\vspace*{-5mm}
\caption{The CIAO program modelling the journey optimization problem.} \label{fig:CiaoJourneys}
\end{figure}

We have a set of facts modelling the user's appointments and the charging stations.
In particular, for each appointment $A_i$, there is a clause $appointment(L_i, \,_it_S^A, id^A)$, while 
for each charging station we have a clause $chargingStation(CSname,SpotsNum,L).$

Moreover, there are four $journey$ clauses describing the structure of journeys.
The upper two represent the base case, while the other two represent the recursive case.
The first clause models the case where
a journey is simply a path with a cost in terms of energy less than or equal to the
SoC of the EV.
The second clause models the case where the SoC of the EV is not enough to do any path and
so a charging event, incrementing the energy level, must be scheduled.
The third $journey$ clause represents the case where a journey is a path
with a cost in terms of energy less than or equal to the
SoC of the EV, plus another journey.
Finally, the last clause models the recursive case where a charging event is needed.
In all cases we check that the paths allow the user to arrive in time.

The head of the journey clauses has the shape
$journey(L_L,L_P,L_{ChEv},[C_T,C_E],SoC)$, where
$L_L$ is the list of the locations of the appointments,
$L_P$ is the list needed to remember, at the end, all the paths 
of the journey in the correct ordering,
$L_{ChEv}$ is the list needed to remember all the charging events needed
to complete the journey,
$[C_T,C_E]$ represents the cost of the journey in terms of time and energy, and 
finally, $SoC$ represents the current energy level of the EV.  

To make the program as readable as possible, we omit
the predicates $newSoC$ and $timeSum$, useful to respectively
compute the new energy level of the EV after a charging event
and the arriving time of the user to an appointment.

The $plus$ clauses are useful to model the soft framework and they are very similar to
the ones of the trip level problem. The only difference is that 
here we have to consider the charging events. Moreover, note that we reuse the $times$
predicate defined in the CIAO program in Fig. \ref{fig:CiaoPaths}.  

The $journeys$ clause collects all
the journeys through a set of locations (the ones of the user's appointments) and returns 
the best solutions chosen with the help of the $plus$ predicate.
So, if we want to know the best journeys, in the graph of Fig. \ref{fig:graph}, 
through the locations where the user has the appointments,
with an EV having an energy level equal to $10$,
we have to perform the CIAO Prolog query $journeys([p,r,t], 10, BestJourneys)$,
where $p,r,t$ are the locations of the appointments and the $BestJourneys$ variable will be instantiated with the 
list containing all the non-dominated journeys. In particular, for each of them,
the list will contain the sequence of the
paths of the journey, the total cost of the
journey in terms of time and energy, and the list of the charging events, each of them described by
the name of the charging station and its location.
The output of the CIAO program for this query is shown in Fig. \ref{fig:jRes}. 
 
\begin{figure}[t]
\begin{center}
\begin{verbatim}
Ciao 1.14.2-13646: Mon Aug 15 10:49:59 2011

?- journeys([p,r,t],10,BestJourneys).

BestJourneys = [
                 [[[p,r],[r,s,t]],2+(3+1),7+(3+1),[[r,csr1]]],
                 [[[p,r],[r,q,t]],2+(1+2),7+(1+4),[[r,csr1]]],
                 [[[p,q,r],[r,s,t]],2+1+(3+1),4+1+(3+1),[]],
                 [[[p,q,r],[r,q,t]],2+1+(1+2),4+1+(1+4),[]]
               ] ?.
no
?-
\end{verbatim}
\end{center}
\vspace*{-5mm}
\caption{The output for the query $journeys([p,r,t],10,BestJourneys)$.} \label{fig:jRes}
\end{figure}

\section{Conclusion} \label{sec:concl}

In this paper we proposed the SCLP framework 
as a high-level declarative, executable specification notation 
 to model in a natural way 
some aspects of the e-mobility optimization
problem \cite{VWPaper}, consisting in coordinating electric vehicles
in order to overcome both energetic and temporal constraints.
In particular, we considered the trip and journey optimization sub-problems, consisting
in finding respectively the energy- and time-optimal route from one destination to another one,
and the optimal sequence of coupled trips, in terms of the same criteria,
guaranteeing that the user reaches each appointment in time.
For both the optimization problems, we provided an SCLP program in CIAO Prolog, by explicitly implementing the 
soft framework, that is, the additive and the multiplicative operations of the chosen semiring.
The former is a slight variant of the CIAO program proposed in \cite[Section 4.4]{DBLP:journals/tocl/BistarelliMRS10} 
to specify the multicriteria 
version of the shortest path problem.
With respect to the program proposed there, here we implemented a different semiring,
(also proposed in \cite{DBLP:journals/tocl/BistarelliMRS10}), i.e., 
the one based on the Hoare Power Domain operator,
which allowed us to obtain only the best 
(i.e. non-dominated) routes in terms of time and energy consumption.
We thus provided an implementation of the two operations of this semiring, by defining two predicates
modelling them.
The SCLP program modelling the journey optimization problem then uses the
trip optimization problem results as inputs. It is also  
based on the same semiring that, in this case, allowed us to find the best journeys in terms of the two cost criteria.

As said above, the soft framework is explicitly implemented into each CIAO
program: there is for example a different $plus$ predicate in each optimization program we have proposed.
However, it would be interesting to study a general way to embed the soft framework 
in Ciao Prolog. Trivially, one could provide a library offering a more general implementation of
the operations of the semiring of each type of problem. 
Most interestingly, one could instead think to provide a meta-level implementing more efficiently the soft framework.

Differently from our solution, which allows us to obtain the set of all the optimal journeys, 
in the mathematical model proposed in \cite{VWPaper} a form of approximation is introduced, by considering
an aggregated cost function to be optimized.
Their goal is indeed to minimize this cost function, which
considers different cost criteria: besides the travel time and the consumed energy, 
they also take into account
the charging cost, the number of charging events, etc.
In modelling the problem, here we introduced a simplification by considering just the two main cost criteria,
that allows us a slender presentation of the work. However, it is obvious that the SCLP programs 
can be easily modified to also take into account several other cost criteria.
On the other side, we preferred not to introduce any approximation of the solution, by
instead returning all optimal journeys considered 
equivalently feasible. However, 
since the use
of partially ordered structures, as in our case, can in general lead 
to a potentially exponential number of undominated solutions,
sometimes it becomes crucial to keep the
number of configurations as low as possible through some form of approximation
allowing us to adopt a total order. In this case, the right
solution could be to adopt a function that composes all the criteria in a single
one and then to choose the best tuple of costs according to the total ordering that 
the function induces
\cite[Section 6.1]{DBLP:journals/tocl/BistarelliMRS10}.

As said above, our aim is mainly to propose the SCLP framework as 
an expressive and natural specification language to model optimization problems.
We indeed think that not only problems representing an extension of the one treated here
can be modelled by adapting the solution we presented easily enough,
but that in general our approach can be followed to model hierarchical optimization problems.
All the SCLP programs we proposed are effective only when data of small size are considered. 
We are indeed conscious that the proposed encodings
cannot be used to really solve the problem on practical cases, but on the other side, 
we think that CIAO represents a powerful system programming environment allowing us 
not only to write declarative specifications but also to reason about them.

It is therefore clear that
here we do not take care of the performances of the proposed programs and that 
our aim is not to compare the 
performance with existing algorithms solving these problems.
We indeed leave as future work the study of how to improve the performance of our programs.
In \cite[Section 8]{DBLP:journals/tocl/BistarelliMRS10}, the authors show some possible 
solutions that could be used towards this end, such as tabling
and branch-and-bound techniques (implementable for example in ECLiPSe \cite{DBLP:books/daglib/0018272}).
We however would also like to study how our programs can take advantage of the use of dynamic programming 
techniques based, for example, on the perfect relaxation algorithm for CSPs \cite{DBLP:conf/iclp/MontanariR91}.

Finally, from a theoretical point of view, as future work, we plan to propose a more 
general framework based on named semiring, allowing us to give a unifying 
presentation of the SCSP and SCLP frameworks, providing an explicit handling of the names.
 

\appendix

\bibliographystyle{splncs03}
\bibliography{biblio}

\end{document}